\documentclass[runningheads]{llncs}
\usepackage{graphicx}
\usepackage{amsmath,amssymb} 
\usepackage{color}
\usepackage[width=122mm,left=12mm,paperwidth=146mm,height=193mm,top=12mm,paperheight=217mm]{geometry}

\usepackage{subfig}

\usepackage{xspace} 
\makeatletter
\DeclareRobustCommand\onedot{\futurelet\@let@token\@onedot}
\def\@onedot{\ifx\@let@token.\else.\null\fi\xspace}

\def\eg{{e.g}\onedot} 
\def\ie{{i.e}\onedot}

\makeatother
\newcommand{\rot}{\rotatebox{90}}
\newcommand{\METHODnospace}{SEC}
\newcommand{\METHOD}{\METHODnospace\xspace}

\begin{document}
\pagestyle{headings}
\mainmatter

\title{Seed, Expand and Constrain: Three Principles for Weakly-Supervised Image Segmentation} 

\titlerunning{Three Principles for Weakly-Supervised Image Segmentation}

\authorrunning{Alexander Kolesnikov and Christoph H. Lampert}

\author{Alexander Kolesnikov \and Christoph H. Lampert}


\institute{IST Austria \\
           \email{\{akolesnikov,chl\}@ist.ac.at}}

\maketitle

\begin{abstract}
We introduce a new loss function for the weakly-supervised training 
of semantic image segmentation models based on three guiding principles:
to \emph{seed} with weak localization cues, to \emph{expand} objects based 
on the information about which classes can occur in an image, and to \emph{constrain} 
the segmentations to coincide with object boundaries.
We show experimentally that training a deep convolutional neural network 
using the proposed loss function leads to substantially better segmentations 
than previous state-of-the-art methods on the challenging PASCAL VOC~2012 dataset.
We furthermore give insight into the working mechanism of our method by a 
detailed experimental study that illustrates how the segmentation quality 
is affected by each term of the proposed loss function as well as their 
combinations. 
\end{abstract}

\section{Introduction}

Computer vision research has recently made tremendous progress.
Many challenging vision tasks can now be solved with high accuracy,
assuming that sufficiently much annotated data is available for training.
Unfortunately, collecting large labeled datasets is time consuming and typically requires substantial financial investments.
Therefore, the creation of training data has become a bottleneck for the further development of computer vision methods.
Unlabeled visual data, however, can be collected in large amounts in a relatively fast and cheap manner.
%
%
Therefore, a promising direction in the computer vision research is to develop methods that can learn from 
unlabeled or partially labeled data. 

In this paper we focus on the task of semantic image segmentation.
Image segmentation is a prominent example of an important vision task, for which creating annotations is especially costly:
as reported in \cite{russakovsky2015imagenet,Bearman15}, manually producing segmentation masks requires several worker-minutes per image.
Therefore, a large body of previous research studies how to train segmentation models from weaker forms of annotation.

%
%
A particularly appealing setting is to learn image segmentation models using training sets with only per-image labels, 
as this form of weak supervision can be collected very efficiently. 
%
However, there is currently still a large performance gap between models trained from per-image labels and models trained from full segmentations masks.
%
In this paper we demonstrate that this gap can be substantially reduced compared to the previous state-of-the-art techniques.

We propose a new composite loss function for training convolutional neural networks for
the task of weakly-supervised image segmentation. 
Our approach relies on the following three insights: 
\begin{itemize}
        \item Image classification neural networks, such as AlexNet~\cite{krizhevsky2012imagenet} or VGG~\cite{simonyan2014very}, can be used to generate reliable object localization cues (\textbf{seeds}), 
        but fail to predict the exact spatial extent of the objects. 
              We incorporate this aspect by using a \textbf{seeding loss} that encourages a segmentation network to match localization cues but that is agnostic about the rest of the image.
        \item To train a segmentation network from per-image annotation, a global pooling layer can be used that aggregates segmentation masks into 
        image-level label scores. 
        The choice of this layer has large impact on the quality of segmentations. For example, max-pooling tends to underestimate the size of 
        objects while average-pooling tends to overestimate it~\cite{pinheiro2015image}. 
              We propose a \textbf{global weighted rank pooling} that is leveraged by \textbf{expansion loss} to expand the object seeds to 
              regions of a reasonable size. 
              It generalizes max-pooling and average pooling and outperforms them in our empirical study.
%

        \item Networks trained from image-level labels rarely capture the precise boundaries of objects in an image.
              Postprocessing by fully-connected conditional random fields (CRF) at test time is often insufficient 
              to overcome this effect, because once the networks have been trained they tend to be confident 
              even about misclassified regions. 
              We propose a new \textbf{constrain-to-boundary loss} that alleviates the problem of imprecise boundaries already at training time. 
              It strives to \textbf{constrain} predicted segmentation masks to respect low-level image information,
              in particular object boundaries.
\end{itemize}

We name our approach \textbf{\METHODnospace}, as it is based on three principles: \textbf{S}eed, \textbf{E}xpand and \textbf{C}onstrain.
We formally define and discuss the individual components of the \METHOD loss function in Section~\ref{sec:method}.
In Section~\ref{sec:experiments} we experimentally evaluate it on the PASCAL VOC 2012 image segmentation benchmark, showing that it  
substantially outperforms the previous state-of-the-art techniques under the same experimental settings.
%
%
We also provide further insight by discussing and evaluating the effect of each of 
our contributions separately through additional experiments. 

\section{Related work}
Semantic image segmentation, \ie assigning a semantic class label to each 
pixel of an image, is a topic of relatively recent interest in computer 
vision research, as it required the availability of modern machine learning 
techniques, such as discriminative classifiers~\cite{shotton2006textonboost,carreira2012cpmc} 
or probabilistic graphical models~\cite{rabinovich2007objects,nowozin2010parameter}. 
As the creation of fully annotated training data poses a major bottleneck to 
the further improvement of these systems, weakly supervised training methods 
were soon proposed in order to save annotation effort.
In particular, competitive methods were developed that only require partial 
segmentations~\cite{verbeek2008scene,he2009learning} or object bounding 
boxes~\cite{liu2012weakly,zhu2014learning,Dai2015ICCV} as   
training data.

A remaining challenge is, however, to learn segmentation models from just 
image-level labels~\cite{vasconcelos2006weakly,verbeek2007region}. 
Existing approaches fall into three broad categories. 
\emph{Graph-based models} infer labels for segments or superpixels based 
on their similarity within or between images~\cite{zhang2013probabilistic,zhang2014probabilistic,zhang2014representative,Xie2014,pourian2015weakly}.
Variants of \emph{multiple instance learning}~\cite{andrews2002support} 
train with a per-image loss function, while internally maintaining a spatial 
representation of the image that can be used to produce segmentation 
masks~\cite{vezhnevets2010towards,vezhnevets2011weakly,vezhnevets2012weakly}.
Methods in the tradition of \emph{self-training}~\cite{scudder1965probability} 
train a fully-supervised model 
but create the necessary pixel-level annotation using the model itself 
in an EM-like procedure~\cite{xu_cvpr2014,xu2015learning,zhang2015weakly}.
Our \METHOD approach contains aspects of the latter two approaches, as it makes 
use of a per-image loss as well as per-pixel loss terms.

In terms of segmentation quality, currently only methods based on deep 
convolutional networks~\cite{krizhevsky2012imagenet,simonyan2014very} 
are strong enough to tackle segmentation datasets of difficulty similar 
to what fully-supervised methods can handle, such as the PASCAL VOC~2012~\cite{everingham2010pascal},
which we make use of in this work.  
In particular, \emph{MIL-FCN}~\cite{pathak2014fully}, \emph{MIL-ILP}~\cite{pinheiro2015image}
and the approaches of~\cite{Bearman15,krapac2016weakly} leverage deep networks in a multiple instance 
learning setting, differing mainly in their pooling strategies, \ie how they convert their 
internal spatial representation to per-image labels.
\emph{EM-Adapt}~\cite{papandreou2015weakly} and \emph{CCNN}~\cite{pathak2015constrained} 
rely on the self-training framework and differ in how they enforce the consistency 
between the per-image annotation and the predicted segmentation masks. 
\emph{SN\_B}~\cite{wei2016learning} adds additional steps for creating and 
combining multiple object proposals.
As far as possible, we provide an experimental comparison to these methods in 
Section~\ref{sec:experiments}.  

\section{Weakly supervised segmentation from image-level labels}\label{sec:method}

In this section we present a technical description of our approach.
We denote the space of images by $\mathcal{X}$. 
For any image $X \in \mathcal{X}$, a segmentation mask $Y$ is a collection, $(y_1, \dots, y_n)$, of semantic labels
at $n$ spatial locations. 
The semantic labels belong to a set $\mathcal{C} = \mathcal{C}^\prime \cup \{c^{\mathrm{bg}}\}$ of size $k$,
where $\mathcal{C}^\prime$ is a set of all foreground labels and $c^{\mathrm{bg}}$ is a background label.
We assume that the training data, $\mathcal{D} = \{(X_i, T_i)\}_{i=1}^{N}$, consists of $N$
images, $X_i \in \mathcal{X}$, where each image is weakly annotated by a set, $T_i \subset \mathcal{C}^\prime$, of foreground labels that 
occur in the image.
Our goal is to train a deep convolutional neural network $f(X; \theta)$, parameterized by $\theta$,
that models the conditional probability of observing any label $c \in \mathcal{C}$ at 
any location $u \in \{1, 2, \dots, n\}$, \ie $f_{u,c}(X; \theta) = p(y_u = c | X)$. 
For brevity we will often omit the parameters $\theta$ in our notation and write $f(X; \theta)$ simply as $f(X)$.

\subsection{The \METHOD loss for weakly supervised image segmentation}

\begin{figure}[t]
        \center
        \includegraphics[width=1.0\textwidth]{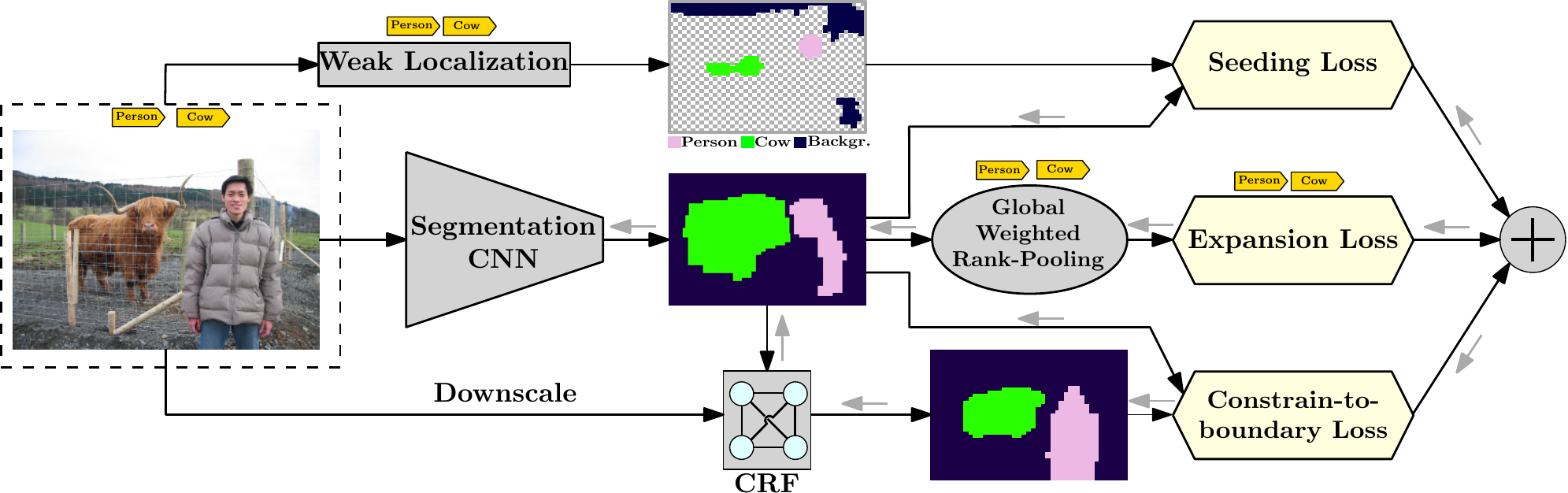}
	\caption{A schematic illustration of \METHOD that is based on 
    minimizing a composite loss function consisting of three terms: \emph{seeding loss}, 
    \emph{expansion loss} and \emph{constrain-to-boundary loss}. See Section~\ref{sec:method} for details. } 
	\label{fig:method}
\end{figure}

Our approach for learning the parameters, $\theta$, of the segmentation neural network
relies on minimizing a loss function that has three terms.
The first term, $L_\textrm{seed}$, provides localization hints to the network, 
%
%
the second term, $L_\textrm{expand}$, penalizes the network
for predicting segmentation masks with too small or wrong objects,
and the third term, $L_\textrm{constrain}$, encourages segmentations that 
respect the spatial and color structure of the images. 
Overall, we propose to solve the following optimization problem for parameter learning:

\begin{align}
        \min\limits_{\theta} \!\!\! \sum\limits_{(X, T) \in \mathcal{D}} \!\!\!\!\! \left[ L_\textrm{seed}(f(X; \theta), T) + L_\textrm{expand}(f(X; \theta), T)
	                                                                               + L_\textrm{constrain}(X, f(X; \theta)) \right].
\end{align}

In the rest of this section we explain each loss term in detail.
A schematic overview of the setup can be found in Figure~\ref{fig:method}.

\subsubsection{Seeding loss with localization cues.}\label{sec:semi}

\begin{figure}[t]
        \center
        \includegraphics[width=0.91\textwidth]{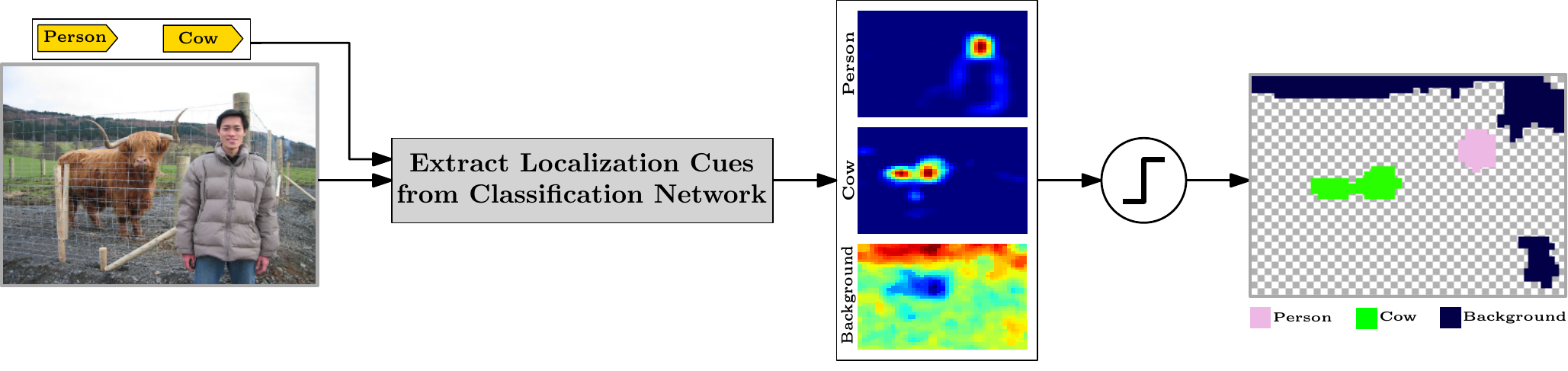}
        \caption{The schematic illustration of the weak localization procedure.}
	\label{fig:local}
\end{figure}

Image-level labels do not explicitly provide any information about the position of semantic objects 
in an image.
Nevertheless, as was noted in many recent research papers~\cite{oquab2015object,zhou2015cnnlocalization,simonyan2013deep,bazzani2016self},
deep image classification networks that were trained just from image-level labels, may be successfully employed to retrieve cues 
on object localization.
We call this procedure \emph{weak localization} and illustrate it in Figure \ref{fig:local}.

Unfortunately, localization cues typically are not precise enough to be used as full 
and accurate segmentation masks. 
However, these cues can be very useful to guide the weakly-supervised segmentation network.
We propose to use a \emph{seeding loss} to encourage predictions of the neural network
to match only ``landmarks'' given by the weak localization procedure while ignoring the rest 
of the image.
Suppose that $S_c$ is a set of locations that are labeled with class $c$ by the 
weak localization procedure.
Then, the \emph{seeding loss} $L_\textrm{seed}$ has the following form:
\begin{align}
	L_\textrm{seed}(f(X), T, S_c) = -\dfrac{1}{\sum\limits_{c \in T} |S_c|} \sum\limits_{c \in T}
							             \sum\limits_{u \in S_c} \log f_{u, c}(X).
\end{align}

Note that for computing $L_\textrm{seed}$ one needs the weak 
localization sets, $S_c$, 
%
so that many existing techniques from the literature can 
be used, essentially, as \emph{black boxes}. 
In this work, we rely on~\cite{zhou2015cnnlocalization} for weakly localizing foreground classes. 
However, this method does not provide a direct way to select confident background 
regions, therefore we use the gradient-based saliency detection 
method from \cite{simonyan2013deep} for this purpose.
We provide more details on the weak localization procedure in Section~\ref{sec:experiments}.

%

\subsubsection{Expansion loss with global weighted rank pooling.}

To measure if a segmentation mask is consistent with the image-level labels one can 
aggregate segmentation scores into classification scores and apply the standard 
loss function for multi-label image classification.
In the context of weakly-supervised segmentation/detection various techniques 
were used by researches to aggregate score maps into a classification scores. 
The most prominent ones are \emph{global max-poling} (GMP)~\cite{oquab2015object} that assigns 
any class $c$ in any image $X$ a score of $\max\limits_{u \in \{1, \dots, n\}} 
f_{u,c}(X)$ and \emph{global average-pooling}~\cite{zhou2015cnnlocalization} that assigns it 
a score of $\ \frac1n\!\sum\limits_{u = 1}^{n} f_{u,c}(X)$.

Both ways of aggregation have been successfully used in practice. However, they have their own drawbacks.
For classes which are present in an image GMP only encourages the response for a single location to be high,
while GAP encourages all responses to be high.
Therefore, GMP results in a segmentation network that often underestimates the sizes of objects,
while network trained using GAP, in contrast, often overestimates them.
Our experiments in Section~\ref{sec:experiments} support this claim empirically. 

In order to overcome these drawbacks we propose a \emph{global weighted rank-pooling (GWRP)}, a new aggregation technique, which can be seen as a generalization of GMP and GAP.
GWRP computes a weighted average score for each class, where weights are higher 
for more promising locations.
This way it encourages objects to occupy a certain fraction of an image, but, unlike GAP, 
is less prone to overestimating object sizes. 

Formally, let an index set $I^c = \{i_{1}, \dots, i_{n}\}$ define the 
descending order of prediction scores for any class $c \in \mathcal{C}$, \ie 
$f_{i_{1},c}(x) \ge f_{i_{2},c}(x) \ge \dots \ge f_{i_{n},c}(x)$ 
and let $0 < d_c <= 1$ be a decay parameter for class $c$.
Then we define the GWRP classification scores, $G_c(f(X), d_c)$, for an image $X$, as following:
\begin{align}
        G_c(f(X); d_c) = \dfrac{1}{Z(d_c)} \sum\limits_{j=1}^n (d_c)^{j-1} f_{i_j,c}(X), \; \textrm{where} 
        \; Z(d_c) = \sum\limits_{j=1}^n (d_c)^{j-1}.
\end{align}
Note, that for $d_c = 0$ GWRP turns into GMP (adopting the convention that $0^0 = 1$), 
and for $d_c=1$ it is identical to GAP. Therefore, GWRP generalizes both approaches and the decay parameter
can be used to interpolate between the behavior of both extremes.

In principle, the decay parameter could be set individually for each class and each image. 
However, this would need prior knowledge about how large objects of each class typically 
are, which is not available in the weakly supervised setting. 
Therefore, we only distinguish between three groups: for object classes that occur in 
an image we use a decay parameter $d_+$, for object classes that do not occur we use 
$d_-$, and for background we use $d_\textrm{bg}$. 
We will discuss how to choose their values in Section~\ref{sec:experiments}. 


In summary, the \emph{expansion loss} term is 
\begin{align}
        L_\textrm{expand}(f(X), T) = &  \! -\dfrac{1}{|T|} \sum\limits_{c \in T} \log G_c(f(X);\! d_+)  \\
                                     &  \! -\dfrac{1}{|\mathcal{C}^\prime \backslash T|} \! \sum\limits_{c \in \mathcal{C}^\prime \backslash T} \!\! \log (1 - G_c(f(X); d_-))
                                           -\log G_{c^{\mathrm{bg}}} (f(X);\! d_\mathrm{bg}). \notag
\end{align}

\subsubsection{Constrain-to-boundary loss.}\label{sec:KL}

The high level idea of the \emph{constrain-to-boundary loss} is 
to penalize the neural network for producing segmentations
that are discontinuous with respect to spatial and color 
information in the input image.
Thereby, it encourages the network to learn to 
produce segmentation masks that match up with object boundaries. 
%

Specifically, we construct a fully-connected CRF, $Q(X, f(X))$,
as in~\cite{krahenbuhl2011efficient}, with unary potentials given by the logarithm of the probability 
scores predicted by the segmentation network, and pairwise 
potentials of fixed parametric form that depend only on 
the image pixels. We downscale the image $X$, so that it matches the resolution of the segmentation mask, produced by the network.
More details about the choice of the 
CRF parameters are given in Section~\ref{sec:experiments}.
We then define the \emph{constrain-to-boundary loss} as the mean 
KL-divergence between the outputs of the network and 
the outputs of the CRF, \ie:
\begin{align}
        L_\textrm{constrain}(X, f(X)) 
                                   &=\dfrac{1}{n} \sum\limits_{u=1}^{n} \sum\limits_{c \in \mathcal{C}} Q_{u,c}(X, f(X)) \log\frac{Q_{u,c}(X, f(X))}{f_{u,c}(X)}. \label{eq:KL}
\end{align}
This construction achieves the desired effect, since it encourages 
the network output to coincide with the CRF output, which itself 
is known to produce segmentation that respect image boundaries. 
An illustration of this effect can be seen in Figure~\ref{fig:method}.

\subsection{Training}
The proposed network can be trained in an end-to-end way using back-propagation, provided 
that the individual gradients of all layers are available. 
%
%
%
%
For computing gradients of the fully-connected CRF we employ the procedure from~\cite{toyoda2008random},
which was successfully used in the context of semantic image segmentation.
Figure~\ref{fig:method} illustrates the flow of gradients for the backpropagation procedure with gray arrows.

\section{Experiments}\label{sec:experiments}

In this section we validate our proposed loss function experimentally,
including a detailed study of the effects of its different terms.




\subsection{Experimental setup}

\textbf{Dataset and evaluation metric.} We evaluate our method on the PASCAL VOC 2012 image 
segmentation benchmark, which has 21 semantic classes, including background~\cite{everingham2010pascal}.
The dataset images are split into three parts: training (\emph{train}, 1464 images), validation (\emph{val}, 1449 images)
and testing (\emph{test}, 1456 images).
Following the common practice we augment the training part by additional images from~\cite{BharathICCV2011}.
The resulting \emph{trainaug} set has 10,582 weakly annotated images that we use to train our models.
We compare our approach with other approaches on both \emph{val} and \emph{test} parts.
For the \emph{val} part, ground truth segmentation masks are available, so we can evaluate 
results of different experiments. We therefore use this data also to provide a 
detailed study of the influence of the different components in our approach.
The ground truth segmentation masks for the \emph{test} part are not publicly available, 
so we use the official PASCAL VOC evaluation server to obtain quantitative results. 
As evaluation measure we use the standard PASCAL VOC~2012 segmentation metric: mean intersection-over-union (mIoU).

\textbf{Segmentation network.}
As a particular choice for the segmentation architecture, in this paper we use \emph{DeepLab-CRF-LargeFOV} from \cite{chen2014semantic}, 
which is a slightly modified version of the 16-layer VGG network \cite{simonyan2014very}.
%
The network has inputs of size 321x321 and produces segmentation masks of size 41x41, 
see~\cite{chen2014semantic} for more details on the architecture.
We initialize the weights for the last (prediction) layer randomly from a normal distribution with 
mean 0 and variance 0.01. All other convolutional layers are initialized from the publicly available VGG model \cite{simonyan2014very}. 
Note, that in principle, our loss function can be combined with any deep convolutional neural network.

\textbf{Localization networks.}
The localization networks for the foreground classes and the background class are also derived from the standard VGG architecture.
In order to improve the localization performance, we finetune these networks for solving a multilabel classification problem on the \emph{trainaug} data.
Due to space limitations we provide exact details on these networks and optimization parameters in the supplementary material.

Note, that in order to reduce the computational effort and memory consumption required for training \METHOD it is possible to precompute the localization cues.
If precomputed cues are available \METHOD imposes no additional overhead for evaluating and storing the localization networks at training time.

\textbf{Optimization.} For training the network we use the batched stochastic gradient descent (SGD) 
with parameters used successfully in \cite{chen2014semantic}.
We run SGD for 8000 iterations, the batch size is 15 (reduced from 30 to allow simultaneous training of two networks), 
the dropout rate is 0.5 and the weight decay parameter is 0.0005. 
The initial learning rate is 0.001 and it is decreased by a factor of 10 every 2000 iterations.
Overall, training on a \texttt{GeForce TITAN-X} GPU takes 7-8 hours, which is comparable to training times of other models, reported, \eg, in~\cite{papandreou2015weakly,pathak2015constrained}.

\textbf{Decay parameters.}
The GWRP aggregation requires specifying the decay parameters, $d_-$, $d_+$ and $d_{\mathrm{bg}}$, 
that control the weights for aggregating the scores produced by the network.
Inspired by the previous research~\cite{papandreou2015weakly,pathak2015constrained} we do so using the following rules-of-thumb
that express prior beliefs about natural images:
\begin{itemize}
 \item for semantic classes that are not present in the image we want to predict as few pixels as possible. 
 Therefore, we set $d_-=0$, which corresponds to GMP.
 \item for semantic classes that are present in the image we suggest that the top 10\% scores represent 50\% of the overall aggregated score.
       For our 41x41 masks this roughly corresponds to $d_+=0.996$.
 \item for the background we suggest that the top 30\% scores represent 50\% of the overall aggregated score, resulting in $d_{\mathrm{bg}} = 0.999$.
\end{itemize}

\begin{figure}[t]
    \center
    \includegraphics[width=1.0\textwidth]{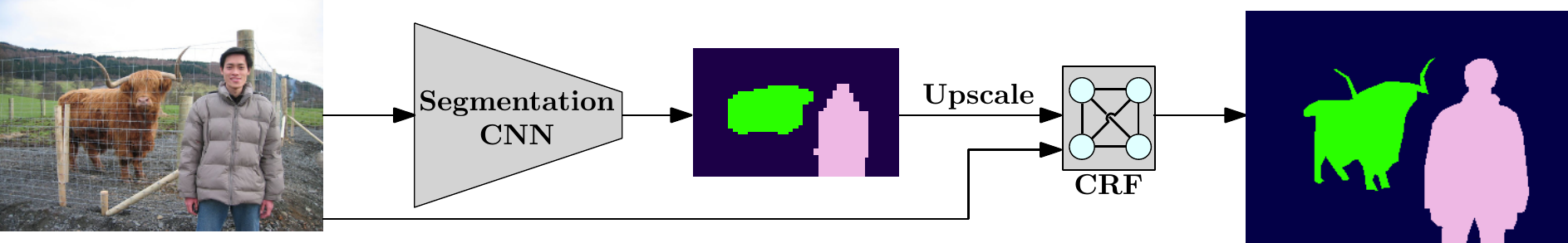}
	\caption{The schematic illustration of our approach at test time.}
	\label{fig:method-test}
\end{figure}

\textbf{Fully-connected CRF at training time}.
In order to enforce the segmentation network to respect the boundaries of objects 
already at training time we use a fully-connected CRF~\cite{krahenbuhl2011efficient}. 
As parameters for the pairwise interactions, we use the default values from the authors'
public implementation, except that we multiply all spatial  
distance terms by $12$ to reflect the fact that we downscaled the original image in order to match the size 
of the predicted segmentation mask.

\textbf{Inference at test time.}
Our segmentation neural network is trained to produce probability scores for all classes and locations,
but the spatial resolution of a predicted segmentation mask is lower than the original image.
Thus, we upscale the predicted segmentation mask to match the size of the input image, and 
then apply a fully-connected CRF~\cite{krahenbuhl2011efficient} to refine the segmentation.
This is a common practice, which was previously employed, \eg, in~\cite{papandreou2015weakly,pathak2015constrained,chen2014semantic}.
Figure~\ref{fig:method-test} shows a schematic illustration of our inference procedure at test time.

\textbf{Reproducibility.}
In our experiments we rely on the \emph{caffe} deep learning framework~\cite{jia2014caffe} 
in combination with a \emph{python} implementation of the \METHOD loss. 
The code and pretrained models are publicly available\footnote{\url{https://github.com/kolesman/SEC}}.
%

\subsection{Results}



\begin{table}\centering
\caption{Results on PASCAL VOC~2012 (mIoU in \%) 
for weakly-supervised semantic segmentation with only per-image labels.}\label{tab:results}
\begin{tabular}{|c||c|c|c|c|c|c|}\hline
\raisebox{1cm}{\parbox{.12\textwidth}{\centering PASCAL\\VOC~2012\\\emph{val} set}}
          & \rot{\cite{Bearman15} (Img+Obj)}  & \rot{\cite{KimH16} (stage1)}	
           & \rot{EM-Adapt}\rot{(re-impl. of~\cite{pathak2015constrained})}	& \rot{CCNN~\cite{pathak2015constrained}}   
           &  \rot{MIL+ILP}\rot{+SP-sppxl${}^{\dag}$~\cite{pinheiro2015image}}  &\rot{SEC (proposed)~}
\\\hline\hline
  background & 	            & 71.7${}^*$	& 67.2	 	& 68.5	 	& 77.2 	& \bf82.4\\
   aeroplane & 	            & 30.7${}^*$	& 29.2	 	& 25.5	 	& 37.3 	& \bf62.9\\
        bike & 	            & \bf30.5${}^*$	& 17.6	 	& 18.0	 	& 18.4 	& 26.4\\
        bird & 	            & 26.3${}^*$	& 28.6	 	& 25.4	 	& 25.4 	& \bf61.6\\
        boat & 	            & 20.0${}^*$	& 22.2	 	& 20.2	 	& \bf28.2 	& 27.6\\
      bottle & 	            & 24.2${}^*$	& 29.6	 	& 36.3	 	& 31.9 	& \bf38.1\\
         bus & 	            & 39.2${}^*$	& 47.0	  	& 46.8	 	& 41.6 	& \bf66.6\\
         car & 	            & 33.7${}^*$	& 44.0	 	& 47.1	 	& 48.1 	& \bf62.7\\
         cat & 	            & 50.2${}^*$	& 44.2	 	& 48.0	 	& 50.7 	& \bf75.2\\
       chair & 	            & 17.1${}^*$	& 14.6	 	& 15.8	 	& 12.7 	& \bf22.1\\
         cow & 	            & 29.7${}^*$	& 35.1	 	& 37.9	 	& 45.7 	& \bf53.5\\
 diningtable & 	            & 22.5${}^*$	& 24.9	 	& 21.0	 	& 14.6 	& \bf28.3\\
         dog & 	            & 41.3${}^*$	& 41.0	 	& 44.5	 	& 50.9 	& \bf65.8\\
       horse & 	            & 35.7${}^*$	& 34.8	 	& 34.5	 	& 44.1 	& \bf57.8\\
   motorbike & 	            & 43.0${}^*$	& 41.6	 	& 46.2	 	& 39.2 	& \bf62.3\\
      person & 	            & 36.0${}^*$	& 32.1	 	& 40.7	 	& 37.9 	& \bf52.5\\
       plant & 	            & 29.0${}^*$	& 24.8	 	& 30.4	 	& 28.3 	& \bf32.5\\
       sheep & 	            & 34.9${}^*$	& 37.4	 	& 36.3	 	& 44.0 	& \bf62.6\\
        sofa & 	            & 23.1${}^*$	& 24.0	 	& 22.2	 	& 19.6 	& \bf32.1\\
       train & 	            & 33.2${}^*$	& 38.1	 	& 38.8	 	& 37.6 	& \bf45.4\\
  tv/monitor & 	            & 33.2${}^*$	& 31.6	 	& 36.9	 	& 35.0 	& \bf45.3\\\hline
     average & 	32.2	& 33.6${}^*$	& 33.8	 	& 35.3	 	& 36.6 	& \bf50.7\\\hline
\end{tabular} 
\begin{tabular}{|c||c|c|c|c|c|}\hline
\raisebox{1cm}{\parbox{.12\textwidth}{\centering~PASCAL\\VOC~2012\\~~\emph{test} set}}
            & \rot{MIL-FCN~\cite{pathak2014fully}}  & \rot{CCNN~\cite{pathak2015constrained}}	
            & \rot{MIL+ILP}\rot{+SP-sppxl${}^{\dag}$~\cite{pinheiro2015image}} & \rot{Region score}\rot{pooling \cite{krapac2016weakly}}
            & \rot{SEC (proposed)~}\\\hline\hline
background	&	& $\approx$71$^{\ddag}$	& 74.7	& $\approx$74$^{\ddag}$ & \bf 83.5                    \\
aeroplane	& 	& 24.2	& 38.8	& 33.1 & \bf 56.4                   \\
bike	& 	& 19.9	& 19.8	& 21.7 & \bf 28.5                  \\
bird	& 	& 26.3	& 27.5	& 27.7 & \bf 64.1                  \\
boat	& 	& 18.6	& 21.7	& 17.7 & \bf 23.6                  \\
bottle	& 	& 38.1	& 32.8	& 38.4 & \bf 46.5                  \\
bus	    &	& 51.7	& 40.0	& 55.8 & \bf 70.6                  \\
car 	& 	& 42.9	& 50.1	& 38.3 & \bf 58.5                  \\
cat	    & 	& 48.2	& 47.1	& 57.9 & \bf 71.3                  \\
chair	& 	& 15.6	& 7.2	& 13.6 & \bf 23.2                  \\
cow	    & 	& 37.2	& 44.8	& 37.4 & \bf 54.0                  \\
diningtable	& 	& 18.3	& 15.8	& \bf 29.2 & 28.0                  \\
dog	    &	& 43.0	& 49.4	& 43.9 & \bf 68.1                  \\
horse 	&	& 38.2	& 47.3	& 39.1 & \bf 62.1                  \\
motorbike& 	& 52.2	& 36.6	& 52.4 & \bf 70.0                  \\
person	&	& 40.0	& 36.4	& 44.4 & \bf 55.0                  \\
plant	&	& 33.8	& 24.3	& 30.2 & \bf 38.4                  \\
sheep	&	& 36.0	& 44.5	& 48.7 & \bf 58.0                  \\
sofa	&	& 21.6	& 21.0	& 26.4 & \bf 39.9                  \\
train	&	& 33.4	& 31.5	& 31.8 & \bf 38.4                  \\
tv/monitor	& 	& 38.3	& 41.3	& 36.3 & \bf 48.3                  \\\hline
average	& 25.7 & 35.6	& 35.8 & 38.0  & \bf 51.7                 \\\hline
\end{tabular}

\hfill{\tiny(${}^\ast$results from unpublished/not peer-reviewed manuscripts, ${}^{\dag}$trained on ImageNet, ${}^{\ddag}$value inferred from average)}

%
\caption{Summary results (mIoU \%) for other methods on PASCAL VOC~2012. Note: the values in this table 
are not directly comparable to Table~\ref{tab:results}, as they were obtained under 
different experimental conditions.}\label{tab:notresults} 
\scalebox{1}{
\begin{tabular}{|l||c|c|l|}\hline
method & \emph{val} & \emph{test} & \text{comments}
\\\hline\hline
DeepLab~\cite{chen2014semantic} & 67.6 & 70.3 & fully supervised training
\\\hline
STC~\cite{WeiLCSCZY15} & 49.8${}^*$ & 51.2${}^*$ & trained on Flickr 
\\
TransferNet~\cite{HongOHL15} & 52.1 & 51.2 & trained on MS~COCO; additional supervision: \\
                            &              &           & from segmentation mask of other classes
\\
\cite{Bearman15} (1Point) & 42.7 & -- & additional supervision: 1 click per class 
\\
\cite{Bearman15} (AllPoints-weighted) & 43.4 & -- & additional supervision: 1 click per instance
\\
\cite{Bearman15} (squiggle) & 49.1 & -- & additional supervision: 1 squiggle per class
\\\hline
EM-Adapt~\cite{papandreou2015weakly} & 38.2 & 39.6 & uses weak labels of  multiple image crops
\\
SN\_B~\cite{wei2016learning} & 41.9 & 43.2 & uses MCG region proposals (see text)
\\
MIP+ILP+SP-seg~\cite{pinheiro2015image} & 42.0 & 40.6 & trained on ImageNet, MCG proposals (see text)
\\
MIL+ILP+SP-bb~\cite{pinheiro2015image} & 37.8 & 37.0 & trained on ImageNet, BING proposals (see text)
\\\hline
\end{tabular}}
\\
\hfill{\scriptsize( ${}^\ast$results from manuscripts that are currently unpublished/not peer-reviewed)}

\end{table}

%
\textbf{Numeric Results.}
Table~\ref{tab:results} compares the performance of our weakly supervised approach with previous 
approaches that are trained in the same setup, \ie using only images from PASCAL VOC~2012 and only 
image-level labels. 
It shows that \METHOD substantially outperforms the previous techniques.
On the \emph{test} data, where the evaluation is performed by an independent 
third party, the PASCAL VOC evaluation server, it achieves 13.5\% higher mean intersection-over-union 
score than the state-of-the-art approaches with new best scores on 20 out of 21 semantic classes.
On the validation data, for which researchers can compute scores themselves, \METHOD improves 
over the state-of-the-art by 14.1\%, and achieves new best scores on 19 out of the 21 classes.

Results of other weakly-supervised methods on PASCAL VOC and the fully-supervised variant of DeepLab 
are summarized in Table~\ref{tab:notresults}. 
We provide these results for reference but emphasize that they should not simply be compared 
to Table~\ref{tab:results}, because the underlying methods were trained on different (and larger) 
training sets or were given additional forms of weak supervision, \eg user clicks. 
Some entries need further explanation in this regard: \cite{papandreou2015weakly} reports results for the EM-Adapt 
model when trained with weak annotation for multiple image crops.
The same model was reimplemented and trained with only per-image supervision in~\cite{pathak2015constrained}, 
so these are the values we report in Table~\ref{tab:results}. 
The results reported for SN\_B~\cite{wei2016learning} and the \emph{seg} variant of the 
MIL+ILP+SP~\cite{pinheiro2015image} are incomparable to others because they were obtained 
with help of \emph{MCG} region proposals~\cite{APBMM2014} that 
were trained in a fully supervised way on PASCAL VOC data.
Similarly, MIL+ILP+SP-\emph{bb} makes use of bounding box proposals generated by the 
BING method~\cite{BingObj2014} that was trained using PASCAL VOC bounding box annotation.

Note that we do include the \emph{sppxl} variant of MIL+ILP+SP in Table~\ref{tab:results}. 
While it is trained on roughly 760.000 images of the ImageNet dataset, we do not consider 
this an unfair advantage compared to our and other methods, because those implicitly benefit 
from ImageNet images as well when using pretrained classification 
networks for initialization. 

\begin{figure}[h!]
    \centering
    \includegraphics[width=0.90\textwidth]{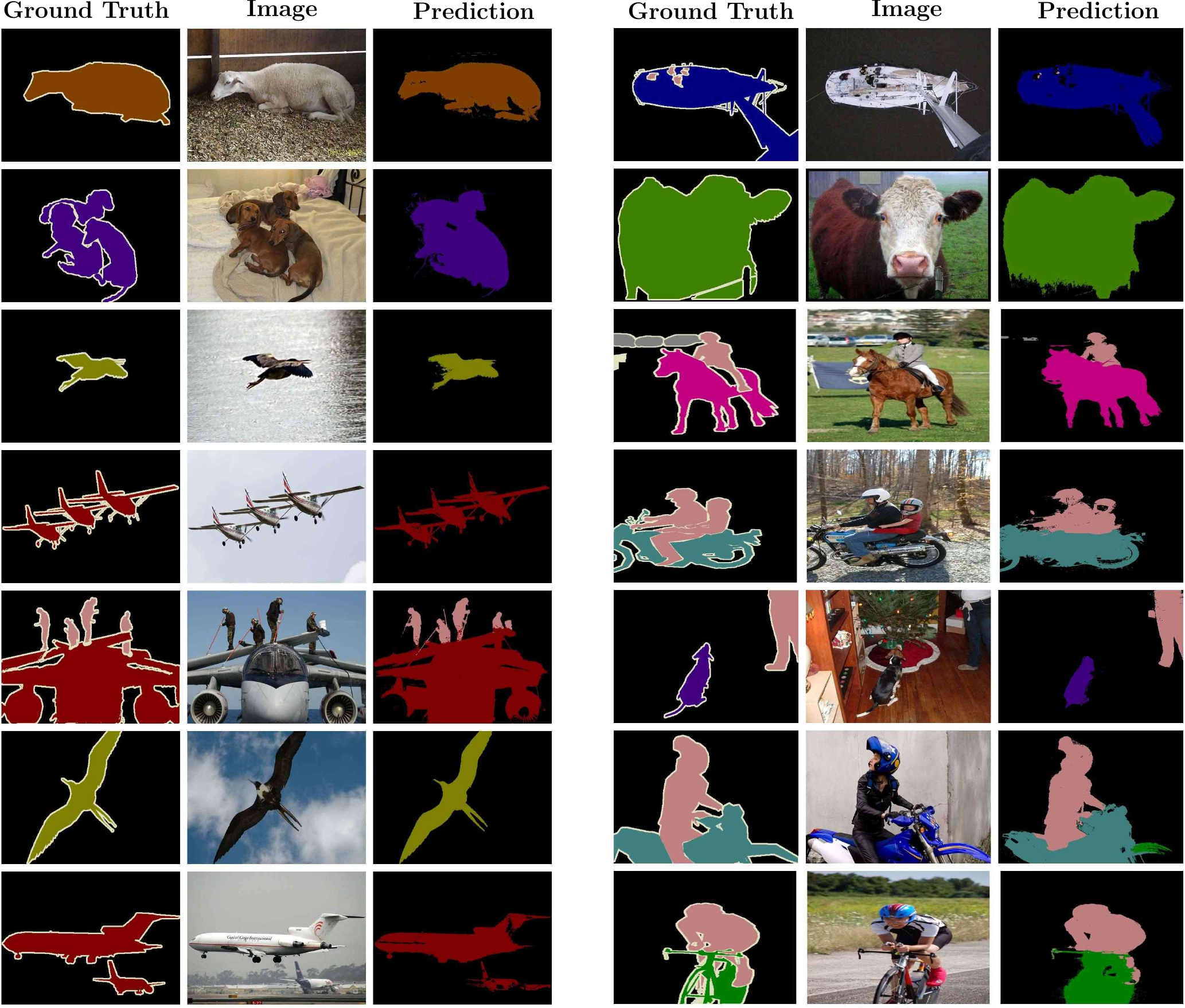}
    \caption{Examples of predicted segmentations (\emph{val} set, successfull cases).}
    \label{fig:results-good}
    \includegraphics[width=0.90\textwidth]{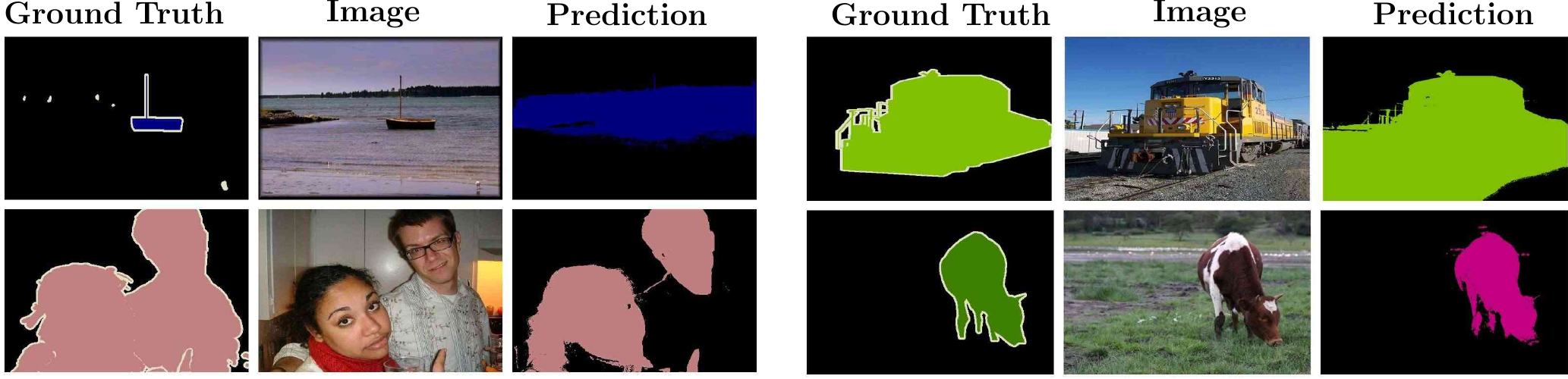}
    \caption{Examples of predicted segmentations (\emph{val} set, failure cases).}
    \label{fig:results-bad}
\end{figure}

\textbf{Qualitative Results.}
%
Figure~\ref{fig:results-good} illustrates typical successful segmentations. 
It shows that our method can produce accurate segmentations even for non-trivial images and recover
fine details of the boundary. 
%
%
Figure~\ref{fig:results-bad} illustrates some failure cases. 
As is typical for weakly-supervised systems, \METHOD has problems segmenting
objects that occur almost always in front of the same background, \eg boats 
on water, or trains on tracks.
We addressed this problem recently in follow-up work~\cite{kolesnikov2016improving}.
%
%
A second failure mode is that object regions can be segmented correctly, 
but assigned wrong class labels. 
This is actually quite rare for \METHOD, which we attribute to the
fact that the DeepLab network has a large field-of-view and therefore 
can make use of the full image when assigning labels. 
Finally, it can also happen that segmentations cover only parts of 
objects.
This is likely due to imperfections of the weak localization cues that tend to 
reliably detect only the most discriminative parts of an object, \eg 
the face of a person. This might not be sufficient to segment the 
complete object, however, especially when objects overlap each other 
or consist of multiple components of very different appearance. 

\subsection{Detailed Discussion}\label{sec:details}

To provide additional insight into the working mechanisms of the
\METHOD loss function, we performed two further sets of experiments
on the \emph{val} data.
%
%
First, we analyze different global pooling strategies, and second, 
we perform an ablation study that illustrates the effect of each 
of the three terms in the proposed loss function visually as well as 
numerically.

\begin{figure}[t]
    \centering
    \includegraphics[width=0.69\textwidth]{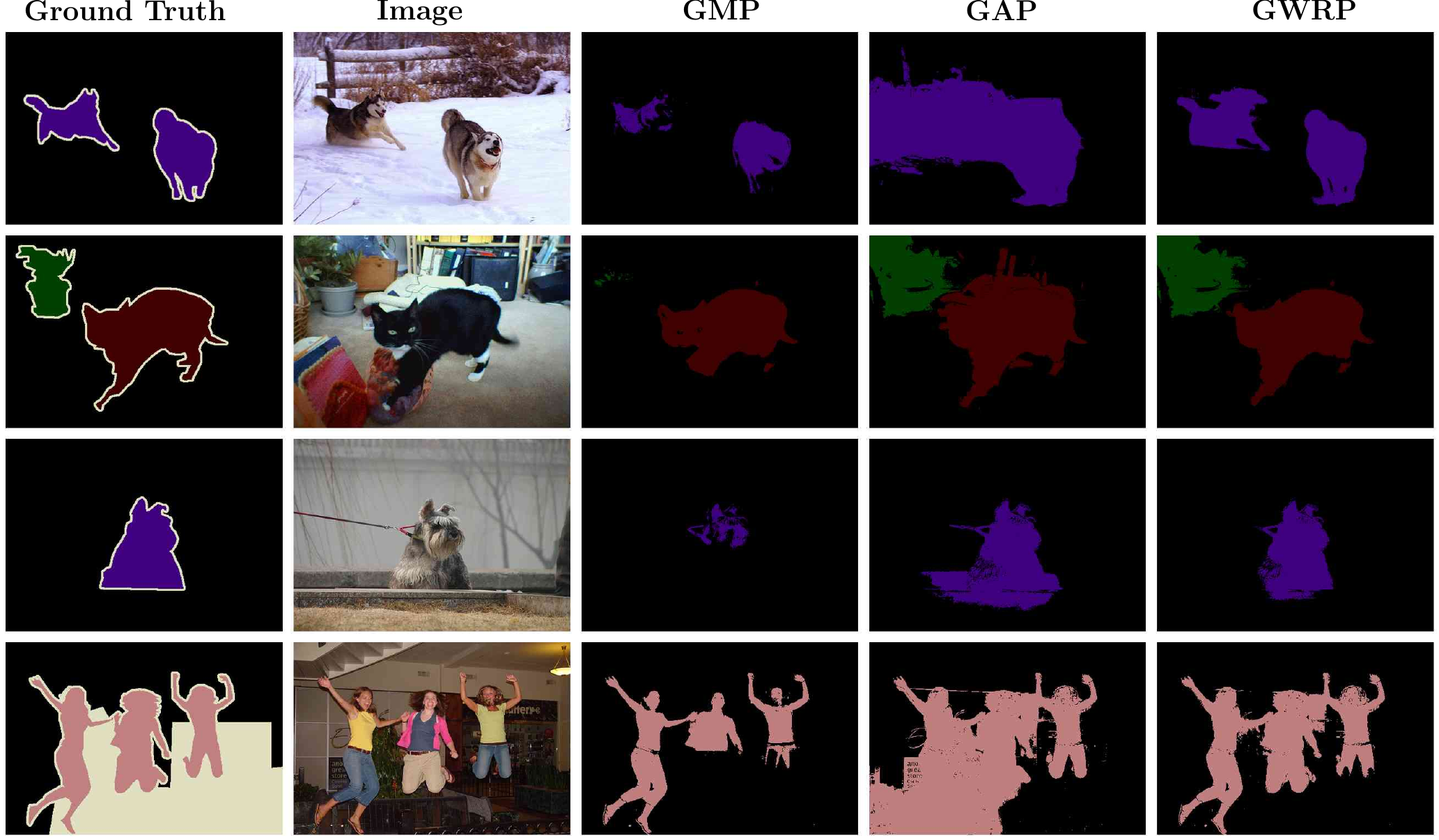}
    \raisebox{6\baselineskip}{
    \begin{tabular}{|c||c|c|}
                \hline
		pooling & fg & mIoU \\
        method      & fraction & \emph{(val)}\\
                \hline
		GMP & 20.4 & 46.5 \\\hline
		GAP & 35.6 & 45.7 \\
                \hline
		GWRP & 25.8 & {50.7} \\ 
                \hline
                \hline
		\colorbox{white}{\parbox{1cm}{\centering ground truth}} & 26.7 & -- \\
                \hline
        \end{tabular}}
    \caption{Results on the \emph{val} set and examples of segmentation masks for models trained with different pooling strategies.}
    \label{fig:pooling}
\end{figure}

\textbf{Effect of global pooling strategies.}
As discussed before, the quality of segmentations depends on which
global pooling strategy is used to convert segmentation mask into 
per-image classification scores. 
To quantify this effect, we train three segmentation networks from weak supervision, 
using either GMP, GAP or GWRP as aggregation methods for classes that are present 
in the image. 
For classes that are not present we always use GMP, \ie we penalize any occurrence of these classes.
%
%
In Figure~\ref{fig:pooling} we demonstrate visual results for every pooling strategy and report two quantities:
the fraction of pixels that are 
predicted to belong to a foreground (fg) class, and the segmentation performance as measured by mean IoU. 
We observe that GWRP outperforms the other method in terms of segmentation quality
and the fractions of predicted foreground pixels supports our earlier hypothesis: 
the model trained with GMP tends to underestimate object sizes, while the model 
trained with with GAP on average overestimates them.
In contrast, the model trained with GWRP, produces segmentations in which 
objects are, on average, close to the correct size\footnote{Note that these experiments 
were done after the network architecture and parameters were fixed. In particular, 
we did not tune the decay parameters for this effect.}.
%

\begin{figure}[t]
    \centering
    \includegraphics[width=0.77\textwidth]{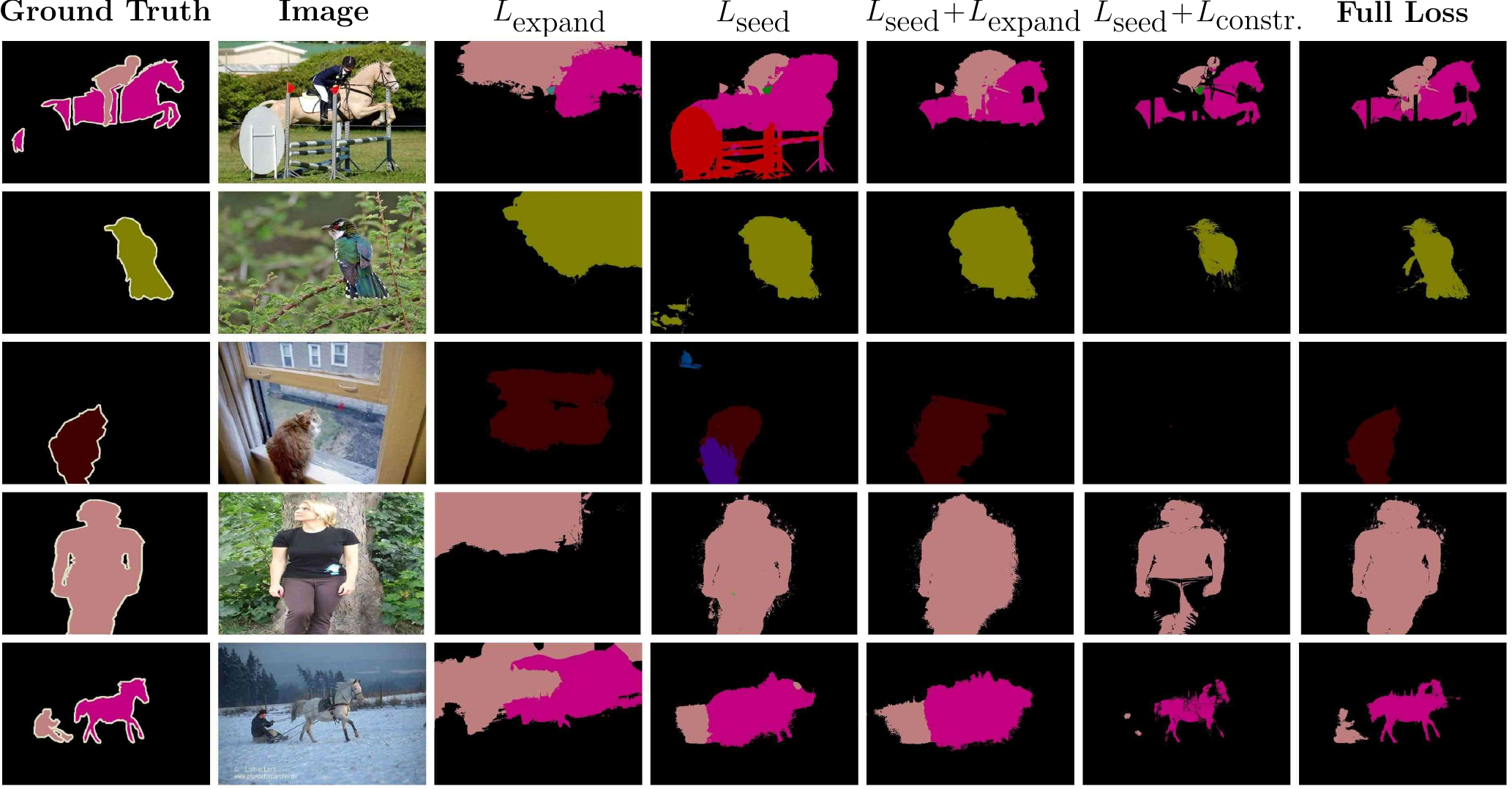}~~
        \raisebox{5.5\baselineskip}{\renewcommand{\arraystretch}{0.9}
        \begin{tabular}{|c|c|}\hline
            loss & mIoU\\
            function & \emph{(val)}\\\hline\hline
            $L_{\textrm{expand}}$ & 20.6\\\hline
            $L_{\textrm{seed}}$ & 45.4 \\\hline
            \colorbox{white}{\parbox{1.3cm}{$L_{\textrm{seed}} +L_{\textrm{expand}}$}}& 44.3\\\hline
            \colorbox{white}{\parbox{1.3cm}{$L_{\textrm{seed}}+L_{\textrm{constrain}}$}}& 50.4 \\\hline
            \parbox{1.3cm}{all terms} & 50.7\\\hline
        \end{tabular}
        }
    \caption{Results on the \emph{val} set and examples of segmentation masks for models trained with different loss functions.}
    \label{fig:ablation}
\end{figure}

\textbf{Effect of the different loss terms.} 
To investigate the contribution of each term in our composite loss function we train 
segmentation networks with loss functions in which different terms of the \METHOD loss 
were omitted.
Figure~\ref{fig:ablation} provides numerical results and
illustrates typical segmentation mistakes that occur when certain loss terms are omitted.
%
Best results are achieved when all three loss terms are present. 
However, the experiments also allow us to draw two interesting additional conclusions
about the interaction between the loss terms.

\textbf{Semi-supervised loss and large field-of-view.}
First, we observe that having $L_\textrm{seed}$ in the loss function is crucial 
to achieve competitive performance.
Without this loss term our segmentation network fails to reflect the 
localization of objects in its predictions, even though the network 
does match the global label statistics rather well. %
See the third column of Figure~\ref{fig:ablation} for the illustration 
of this effect.

\begin{figure}[t]
    \centering
    \includegraphics[width=0.77\textwidth]{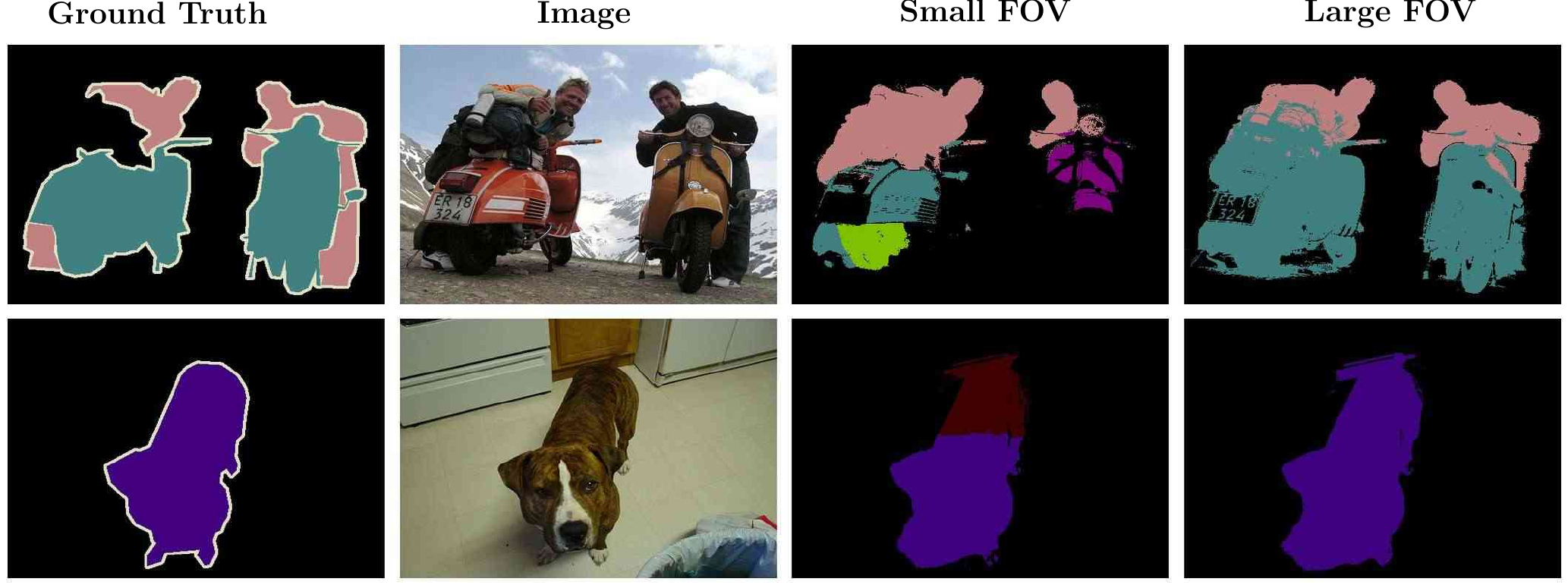}
\quad\raisebox{3.2\baselineskip}{\begin{tabular}{|c|c|}\hline
field & mIoU \\
of view & \emph{(val)} \\\hline
211x211 & 38.1 \\\hline
378x378 & 50.7 \\\hline
\end{tabular}}
    \caption{Results on the \emph{val} set and examples of segmentation masks for models with small or large field-of-views.}
    \label{fig:field}
\end{figure}

We believe that this effect can be explained by the large (378x378) field-of-view (FOV) of the 
segmentation network\footnote{We report
the theoretical fields-of-view inferred from the network architecture.
The empirical field-of-view that is actually used by the network can be smaller~\cite{zhou2014object}.}:
if an object is present in an image, then the majority 
of the predicted scores may be influenced by this object, no matter where 
object is located.
This helps in predicting the right class labels, but can negatively affect
the localization ability. 
Other researchers addressed this problem by explicitly changing the architecture 
of the network in order to reduce its field-of-view~\cite{papandreou2015weakly}. 
However, networks with a small field-of-view are less powerful and often 
fail to recognize which semantic labels are present on an image.
We conduct an additional experiment (see the supplementary material for details) 
that confirm that \METHOD with a small (211x211) field-of-view network performs clearly 
worse than 
with the large (378x378) field-of-view network, see Figure~\ref{fig:field} for numeric results and visual examples.
Thus, we conclude that the seeding loss provides the necessary localization 
guidance that enables the large field-of-view network to still reliably localize 
objects. 
%

\textbf{Effects of the expansion and constrain-to-boundary losses.}
By construction, the constrain-to-boundary loss encourages nearby regions of 
similar color to have the same label. 
However, this is often not enough to turn the weak localization cues 
into segmentation masks that cover a whole object, especially if the 
object consists of visually dissimilar parts, such as people wearing 
clothes of different colors. 
See the sixth column of Figure~\ref{fig:ablation} 
for an illustration of this effect. 

The expansion loss, based on GWRP, suppresses the prediction of classes 
that are not meant to be in the image, and it encourages classes that are in 
the image to have reasonable sizes.
When combined with the seeding loss, the expansion loss 
actually results in a drop in performance. The fifth column of Figure~\ref{fig:ablation} 
shows an explanation of this: objects sizes are generally increased, but 
the additionally predicted regions do not match the image boundaries. 

In combination, the seeding loss provides reliable seed locations,
the expansion loss acts as a force to enlarge the segmentation masks 
to a reasonable size, and the constrain-to-boundary loss constrains the segmentation 
mask to line up with image boundaries, thus integrating low-level image 
information. 
The result are substantially improved segmentation masks as illustrated 
in the last column of Figure~\ref{fig:ablation}.

\section{Conclusion}

We propose a new loss function for training
deep segmentation networks when only image-level labels are available.
We demonstrate that our approach outperforms previous state-of-the-art 
methods by a large margin when used under the same experimental conditions
and provide a detailed ablation study.

We also identify potential directions that may help to further improve 
weakly-supervised segmentation performance.
Our experiments show that knowledge about object sizes can dramatically 
improve the segmentation performance.
\METHOD readily allows incorporating size information through decay parameters, 
but a procedure for estimating object sizes automatically would be desirable. 
A second way to improve the performance would be stronger segmentation priors,
for example about shape or materials. 
This could offer a way to avoid mistakes that are currently typical for 
weakly-supervised segmentation networks, including ours, for example that 
boats are confused with the water in their background.

\textbf{Acknowledgments.} This work was funded by the European Research Council under
the European Union’s Seventh Framework Programme (FP7/2007-2013)/ERC grant agreement no
308036. We gratefully acknowledge the support of NVIDIA Corporation with the
donation of the GPUs used for this research.
We also thank Vittorio Ferrari for helpful feedback.

\bibliographystyle{splncs03}
\bibliography{0842}

\clearpage
\section{Supplementary material}

In this section we provide technical details of
the weak localization procedure (section \ref{sec:semi}) and specify how the network with the small field-of-view was derived (section \ref{sec:details}).

\subsection{Localization networks}

The seeding loss \METHOD relies on weak localization cues.
As was noted in many recent papers~\cite{oquab2015object,zhou2015cnnlocalization,simonyan2013deep,bazzani2016self},
localization cues may be produced by leveraging a deep convolutional neural network that is trained
for solving an image classification task.
We use two different approaches, one for localizing the foreground classes and the other for the background class.

Specifically, for localizing the foreground classes we employ the technique from~\cite{zhou2015cnnlocalization}.
As an underlying classification network we use the standard VGG network, initialized from the publicly available model~\cite{simonyan2014very}.
The VGG architecture is slightly modified in order to make the methodology from~\cite{zhou2015cnnlocalization} applicable.
In particular, we implement the following changes into VGG:
\begin{itemize}
        \item the last two fully-connected layers, \emph{fc}6 and \emph{fc}7, are substituted with randomly initialized convolutional layers, which have 1024 output channels and kernels of size 3.
        \item the output of the last convolutional layer is followed by a global average pooling layer and
              then by a fully-connected prediction layer with 20 outputs (the number of foreground semantic classes in PASCAL VOC)
\end{itemize}
Additionally, in order to increase the spatial resolution of the last convolutional layer of the network, 
we increase the input size to 321x321 and omit the two last max-pooling layers, \emph{pool}4 and \emph{pool}5.
The resulting network is finetuned with multilabel logistic loss on the \emph{trainaug} part
of the PASCAL VOC 2012 dataset (we use the same optimization parameters as in Section~\ref{sec:experiments}).
Then the network is used to provide class-specific localization heat maps.
In order to produce localization cues for every foreground class
we threshold the corresponding heat map by 20\% of its maximum value, as was suggested in~\cite{zhou2015cnnlocalization}.
The resulted localization cues are stacked together into a single weak localization mask, as shown on Figure \ref{fig:local}.
It may happen that localization cues are in conflict, when different classes are assigned to the same location.
We use simple rule to resolve these conflicts: during the stacking process classes that occupy a smaller fraction of an image have priority over classes that occupy a bigger fraction of an image.

For localizing background we rely on the alternative technique from~\cite{simonyan2013deep}.
We also use the VGG as the underlying network.
It is modified to have input resolution of size 321x321 and to have a prediction layer with 20 outputs.
Analogously to \cite{chen2014semantic}, we also change the number of output channels in the fully-connected layers, \emph{fc}6 and \emph{fc}7, from 4096 to 1024.
We finetune the network using the same procedure as for the network for localizing the foreground classes.
Following the gradient-based procedure from \cite{simonyan2013deep},
we utilize the finetuned network to produce class-independent saliency maps.
The saliency maps can be quite noisy, so me smoothen them by the median filter with a window of size $3 \times 3$.
Finally, 10\% of the least salient locations in each image are selected as background cues.

\subsection{Small field-of-view}
The \emph{Deeplab-Large-FOV} neural network achieves particularly wide field-of-view by utilizing convolutions with ``holes'',
which were recently suggested in the context of semantic image segmentation in \cite{papandreou2015weakly}.
In order to derive the closest architecture, but with a small field-of-view, we substitute convolutions with ``holes'' by the standard convolutional layers.
This leads to a measurable drop in the size of the field-of-view: from 378x378 to 211x211.
\begin{figure}[t]
    \center
    \includegraphics[height=0.97\textheight,width=1.0\textwidth]{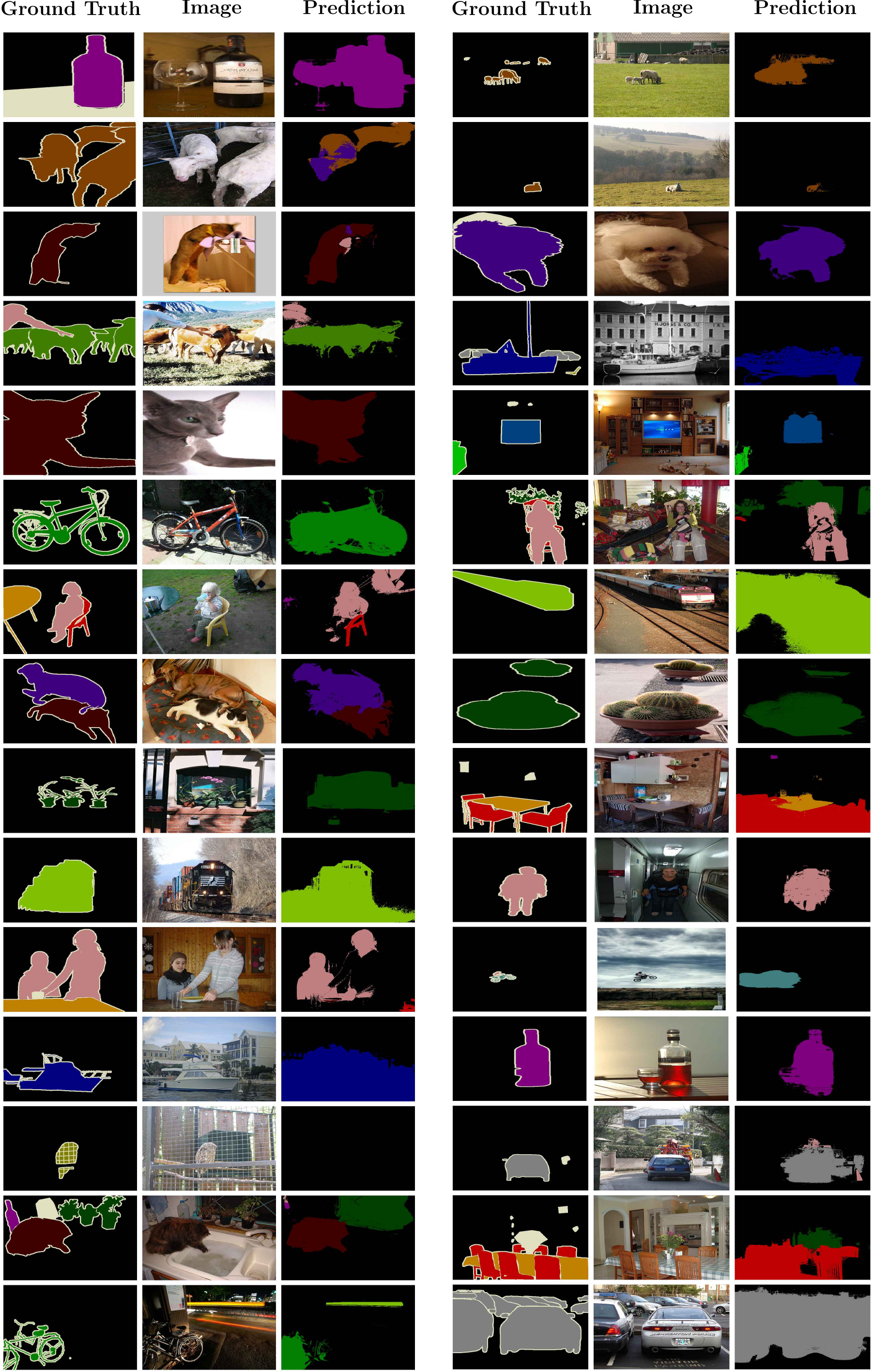}
    \caption{Random examples of segmentations produced by \METHOD (\emph{val} set).} 
	\label{fig:random-1}
\end{figure}

\end{document}